# Automatic Emotion Recognition (AER) System based on Two-Level Ensemble of Lightweight Deep CNN Models

Emad-ul-Haq Qazi[1], Muhammad Hussain[1], Hatim AboAlsamh[1], Ihsan Ullah[2]

[1]Visual Computing Lab, Department of Computer Science, College of Computer and Information Sciences, King Saud University, Riyadh 11543, Saudi Arabia.

[2]Insight Centre for Data Analytics, National University of Ireland, Galway, Ireland.

**Abstract**: Emotions play a crucial role in human interaction, health care and security investigations and monitoring. Automatic emotion recognition (AER) using electroencephalogram (EEG) signals is an effective method for decoding the real emotions, which are independent of body gestures, but it is a challenging problem. Several automatic emotion recognition systems have been proposed, which are based on traditional hand-engineered approaches and their performances are very poor. Motivated by the outstanding performance of deep learning (DL) in many recognition tasks, we introduce an AER system (Deep-AER) based on EEG brain signals using DL. A DL model involves a large number of learnable parameters, and its training needs a large dataset of EEG signals, which is difficult to acquire for AER problem. To overcome this problem, we proposed a lightweight pyramidal one-dimensional convolutional neural network (LP-1D-CNN) model, which involves a small number of learnable parameters. Using LP-1D-CNN, we build a two level ensemble model. In the first level of the ensemble, each channel is scanned incrementally by LP-1D-CNN to generate predictions, which are fused using majority vote. The second level of the ensemble combines the predictions of all channels of an EEG signal using majority vote for detecting the emotion state. We validated the effectiveness and robustness of Deep-AER using DEAP, a benchmark dataset for emotion recognition research. To identify the brain region that has dominant role in AER, we analyzed EEG signals over five brain regions: *FRONT, CENT, PERI, OCCIP* and *ALL*. The results indicate that *FRONT* plays dominant role in AER and over this region, Deep-AER achieved the accuracies of 98.43% and 97.65% for two AER problems, i.e., high valence vs low valence (HV vs LV) and high arousal vs low arousal (HA vs LA), respectively. The comparison reveals that Deep-AER outperforms the state-of-the-art systems with large margin. The Deep-AER system will be helpful in monitoring for health care and security investigations.

## 1. Introduction

Emotion is a psycho-physiological process triggered by conscious and/or unconscious perception of an object or situation and is often associated with temperament, mood, motivation and personality. Emotions play an important role in human health care, communication and security investigations and can be expressed either verbally through emotional statements or by expressing non-verbal cues such as facial expressions, intonation of voice, and body gestures [1]. Emotions effect decision making, mutual interaction and cognitive processes [2]. With the advancement of technology and the understanding of emotions, there are growing opportunities for automatic emotion recognition (AER) systems. There have been many research studies on emotion recognition using different modalities such as facial expressions, speech, text or gestures but these modalities are based on audio and visual observations, which can be easily disguised. The alternative is to use electroencephalogram (EEG) to capture brain signals activated by various types of emotions. EEG is a commonly used neuroimaging technique to analyze neural processes, and from the clinical point of view, it captures the brain activations directly. The EEG brain signals can directly map the brain states, which represent different emotion states and cannot be disguised. As such EEG brain signals form a better modality to detect true emotions.

According to the research in psychology, two major approaches for modeling emotions are: (1) categorical approach and (2) dimensional approach [3]. The categorical approach was pioneered by Darwin et al. [4] and its focus is on basic emotions: happiness, sadness, surprise, disgust, anger and fear [5]. According to the dimensional approach, affective states are not independent, rather, they are related to one another in a systematic manner. In this approach, a model of emotion is characterized by two dimensions i.e. valence and arousal [6-8]. The valence is the degree of aversion or attraction that an individual feels about a specific event or object. It ranges from negative (unhappy) to positive (happy). The arousal is a physiological and psychological state of being reactive to stimuli, ranging from passive (inactive) to active. Dimensional approach is assumed to be better than categorical approach as it describes a larger set of emotions [9]. Therefore, the problem of emotion recognition at high level is to classify HV vs LV and HA vs LA. Recently, a number of research studies employed the EEG brain signals for the classification of HV vs LV and HA vs LA. Most of the research work used machine learning (ML) techniques based on hand-engineered features [10-16], which show poor performance in emotion recognition. The pioneer work by Koelstra et al. [10] extracted power spectrum features from 32 EEG signals and classified them using gaussian naive Bayes classifier into two levels of valence and arousal each. Alazrai et al. [11] also employed EEG signals from DEAP, extracted quadratic time-frequency distribution (QTFD) based features, and used SVM as a classifier. Huang et al. [12] introduced asymmetry spatial pattern (ASP) as a feature extracted from EEG signals, and used naive K-Nearest Neighbor (K-NN), Bayes (NB), and SVM for emotion classification. Similarly, several other studies used hand-engineered based techniques for feature extraction from EEG signals and performed emotion classification using various classifiers such as naive Bayes, SVM, K-NN, LDA, and ANN [13-16]. The maximum reported accuracies for the classification of HV vs LV and HA vs LA are 85.8% and 86.6% on DEAP database, a public domain database for emotion recognition [11]. It indicates that hand-engineered features cannot properly represent the discriminative patterns from EEG signals that are relevant to emotions and the existing methods have not reached the desired level to classify human emotions. It means that the emotion recognition from EEG is still a challenging problem.

Motivated by the outstanding performance of deep learning (DL) in many recognition tasks, we used DL to develop a robust and more effective AER system (Deep-AER) based on EEG brain signals. DL is the state-of-the-art machine learning technique, which automatically learns hierarchy of features and classifies them in an end-to-end fashion [17]. The features extracted by DL models are adapted to the inherent structural patterns of data and due to this reason, they are more discriminative and robust than hand-engineered features [18]. The most effective deep architecture is convolutional neural network (CNN). The various 2D and 3D CNN models such as AlexNet [19], VGG [20], 3DCNN [21], C3D [22] have shown excellent performance in many fields. Recently, 1D CNN models has been successfully used for music generation, epilepsy detection, text understanding, and other time series data [23-27]. The DL being an end-to-end learning approach needs not the design of feature descriptors, the selection of most discriminative features and the adaptation of a suitable classifier [28-31].

As a DL model involves a large number of learnable parameters, its training needs a large dataset of EEG signals, which is difficult to acquire for AER problem. To overcome this issue, we proposed a lightweight pyramidal one-dimensional convolutional neural network (LP-1D-CNN) model, which contains a less number of learnable parameters. An EEG signal consists of a number of channels, where each channel is a 1D signal. To predict emotions from an EEG signal, each channel is to be analyzed. The temporal length of each channel is normally large e.g. the temporal length of each channel in DEAP is 1 minutes, which consists of 8064 samples. For the analysis of each channel if a 1D CNN model is used, its complexity becomes high because of a large input signal and it is prone to overfitting. To overcome this issue, first we segment each channel into small windows, and train one LP-1D-CNN model on these windows; as the size of each input window is small, the complexity of the LP-1D-CNN model is low and it is robust against overfitting. At test time, the predictions of all windows of a channel by the LP-1D-CNN model are fused. Further, the decisions from each channel are composed to predict the emotion state from the EEG signal. In this way, using LP-1D-CNN, we build a two level ensemble model for the classification of emotions.

We validated the effectiveness and robustness of Deep-AER using DEAP, a benchmark dataset for emotion recognition research. We focused on two emotion classification problems: HV vs LV and HA vs LA. To identify the brain region that has dominant role in AER, we analyzed EEG signals over five brain regions: *FRONT, CENT, PERI, OCCIP* and *ALL*. The results indicate that *FRONT* plays dominant role in AER and over this region, Deep-AER achieved the accuracies of 98.43% and 97.65% for HV vs LV and HA vs LA, respectively. It outperforms the state-of-the-art techniques by large margin. The main contributions of this work are: 1) a new lightweight 1D CNN model – LP-1D-CNN model and a data augmentation technique for its training, 2) the Deep-AER system for the classification of emotions from EEG signals based on two-level ensemble of LP-1D-CNN models, 3) the analysis of brain regions to identify the one that plays dominant role in emotion recognition, and 4) a thorough evaluation of the proposed system on the benchmark database DEAP that demonstrates it can be reliably employed for emotion classification in different application scenarios.

The rest of the paper is organized as follows: In Section 2, we present the literature review. Section 3 describes in detail the Deep-AER system framework based on DL. Section 4 presents the experimental protocol and evaluation criteria. Section 5 presents the results and discussion. In the end, Section 6 concludes the paper.

## 2. Literature Review

Many EEG based automatic systems for emotion recognition have emerged in recent years; these systems use different approaches. The categorization of emotions is a classification problem, which involves extraction of discriminatory features from EEG signals and then performing classification. In the following paragraphs, we review the state-of-the-art techniques that have been proposed for emotions classification.

The pioneer work by Koelstra et al. [10] extracted power spectrum density (PSD) features from EEG signals and classified them using gaussian naive Bayes classifier into two levels of valence and arousal each. This method achieved an accuracy of 57.6% for HV vs LV and 62% for HA vs LA on DEAP database, a public domain database for emotion recognition. Alazrai et al. [11] employed EEG signals from DEAP, extracted QTFD-based features, and used SVM as a classifier. The authors performed the classification of valence and arousal into two states (high and low) and achieved the accuracy of 85.8% and 86.6% for HV vs LV and HA vs LA, respectively. Huang et al. [12] introduced asymmetry spatial pattern (ASP) as a feature extracted from EEG signals, and used K-Nearest Neighbor (K-NN), naive Bayes (NB), and SVM for emotion classification. The average accuracies achieved by this method for valence (HV/LV) and arousal (HA/LA) were 66.05% and 82.46%, respectively, on DEAP. Chung and Yoon [32] performed the classification of valence and arousal by using statistical and shallow learning methods like Bayesian classification. The authors extracted the power spectral features from EEG signals and classified them using Bayes classifier. They used the DEAP dataset and divided the valence and arousal into two classes, i.e., HV vs LV and HA vs LA, respectively. They achieved 66.6% and 66.4% accuracy on valence and arousal, respectively. Candra et al. [33] used the discrete wavelet transform (DWT) for extracting time-frequency domain features from EEG signals using DEAP dataset. They computed the entropy of the detail coefficients corresponding to the alpha, beta, and gamma bands and used SVM as a classifier to perform the classification of valence and arousal into two classes (high and low). This method achieved an accuracy of 65.13% for valence (HV/LV) and 65.33% for arousal (HA/LA). In another study, Rozgic et al. [34] developed a method for the classification of EEG signals into two classes, i.e., HV vs LV and HA vs LA by considering the valence and arousal dimensions of DEAP dataset. Firstly, they divided the EEG signals into overlapping segments. Then discriminative features were extracted from each of the EEG segment using the PSD. After features extraction, authors used three different classifiers to classify EEG signals, i.e., SVM, naive Bayes NN (NB-NN) and nearest neighbor (NN) voting. Out of the three classifiers, SVM gave the best classification accuracy of 76.9% and 69.1% for HV vs LV and HA vs LA, respectively. Abeer et al. [35] proposed a method to classify the valence and arousal dimensions into two classes, i.e. HV vs LV and HA vs LA. The authors extracted the PSD and pre-frontal asymmetry features from EEG signals and used deep neural network (DNN) as a classifier. The proposed technique achieved the accuracy of 82% for each of HV vs LV and HA vs LA on DEAP dataset. The method proposed in Zhang et al. [36] extracted power spectral and statistical features from EEG signals using DEAP dataset and classified them using J48 classifier. They divided the data into two classes, HV/LV and HA/LA. The authors used the ontological model for integration and representation of EEG data. This method achieved an accuracy of 75.19% and 81.74% on HV vs LV and HA vs LA, respectively. In another study, Liu et al. [37] proposed an approach based on deep belief networks (DBNs) for the classification of valence and arousal dimensions into two classes (HV/LV and HA/LA) using DEAP dataset. The accuracies obtained using the DBNs approach were 85.2% and 80.5% for HV vs LV and HA vs LA classes, respectively. Atkinson and Campos [38] extracted a set of features from EEG signals in DEAP dataset such as Hjorth parameters and fractal dimension, statistical features and band power for various frequency bands. In this approach, the authors divided the EEG signals into two classes, HV vs LV and HA vs LA and used SVM as a classifier to classify them. They also used the mRMR algorithm for the selection of subset of discriminative features from the set of extracted features. The results reported in [38] showed that the accuracies obtained for HV s LV and HA vs LA classes were 73.1% and 73.0%. In another study, Tripathi et al. [39] extracted the statistical time-domain features from EEG signals and used two types of neural networks, deep NN and convolutional NN as a classifier to discriminate between the EEG signals into two classes, HV vs LV and HA vs LA. The accuracies obtained on DEAP dataset using deep NN in discriminating the HV/LV and HA/LA classes were 75.78% and 73.12%, respectively. Similarly, the accuracies attained in classifying the HV/LV and HA/LA classes using the convolutional NN were 81.4% and 73.3%, respectively. Yin et al. [40] used multiple fusion layer based ensemble classifier of stacked autoencoder (MESAE) to classify EEG signals into HV vs LV and HA vs LA classes. The authors extracted the power spectral and statistical features from EEG signals. The experimental results reported in [40] showed that the accuracies achieved for HV/LV and HA/LA classes were 83.04% and 84.18% on DEAP dataset, respectively. In another study, Zhuang et al. [41] used empirical mode decomposition (EMD) method for emotion recognition. In this approach, EMD based features were extracted from the EEG signals and used SVM as a classifier to discriminate between the HV vs LV and HA vs LA classes. The results showed that the accuracies attained by using this method were 69.1% and 71.9% for HV/LV and HA/LA classes on DEAP dataset, respectively. Li et al. [42] introduced a method for emotion recognition using EEG signals. The authors extracted the nonlinear dynamic domain, frequency-domain and time-domain features from EEG signals and used weighted fusion SVM as a classifier. The EEG signals in DEAP dataset were divided into two classes HV vs LV and HA vs LA. The accuracies obtained in classifying the HV/LV and HA/LA classes were 80.7% and 83.7%, respectively. In another study, Menezes et al.

[43] proposed an approach for emotion recognition based on different combinations of features and different classifiers. The authors extracted the PSD, higher order crossings (HOC) and statistical features from EEG signals and used different classifiers such as random forest and SVM. The results reported in [43] showed that the best accuracies achieved for HV vs LV and HA vs LA classes using SVM classifier were 88.4% and 74.0% on DEAP dataset, respectively.

The overview of the state-of-the-art methods given above indicates that most of the existing methods do not give good performance for emotion recognition. They are based on hand-crafted features, which do not extract the discriminative information from EEG signals well and their performance depends on the tuning of various parameters. These techniques do not generalize well because the hand-engineered features usually are not learned from the data under study and do not encode their structural patterns. In view of the decisive victory of DL over hand-engineered features [17-22], DL can be employed to improve the generalization and accuracy of an emotion recognition system.

## 3. Deep-AER System based on Deep Learning

The proposed AER system is based on two level ensemble of deep LP-ID-CNN models, the architectures of first and second level ensembles are shown in Fig. 1 and 2. The detailed architecture of an LP-ID-CNN mode, its training and testing are given in the following subsections. We represent an EEG signal captured with $C$ electrodes over a time interval with $T$ timestamps $t_1, t_2, \ldots, t_T$ as a $C \times T$ matrix,

$$X = \begin{bmatrix} X_1 \\ X_2 \\ \cdot \\ \cdot \\ \cdot \\ X_c \end{bmatrix} = \begin{bmatrix} x_1(t_1) & x_1(t_2) & \ldots \ldots & x_1(t_T) \\ x_2(t_1) & x_2(t_2) & \ldots \ldots & x_2(t_T) \\ \cdot & \cdot & & \cdot \\ \cdot & \cdot & & \cdot \\ \cdot & \cdot & & \cdot \\ x_c(t_c) & x_c(t_2) & \ldots \ldots & x_c(t_T) \end{bmatrix} \quad (1)$$

where $X_i = [x_i(t_1)\ x_i(t_2) \ldots x_i(t_T)] \in R^T$ is the $i^{th}$ channel of $X$ captured from the $i^{th}$ electrode.

The first level ensemble is designed to take decision about the state of each channel $X_i$. It consists of three main modules: (i) splitting the input channel $X_i$ into $K$ non-overlapping sub-signals using a window of fixed temporal length $T_w$, i.e. $X_i = \{S_1^i\ S_2^i \ldots S_k^i\}$ where $S_k^i = [x_i(t_{j_k})\ x_i(t_{j_k+1}) \ldots x_i(t_{j_k+T_w})] \in R^{T_w}$ and $k = 1, 2, \ldots, K$ (ii) classification of each sub-signal $S_k^i$ with the LP-1D-CNN model $M^i$ corresponding to channel $X_i$, i.e. $O_k^i = M^i(S_k^i)$ where $O_k^i \in \{0, 1\}$ is the predicted label of $S_k^i$, $k = 1, 2, \ldots, K$ and (iii) fusing the predictions of all sub-signals $S_k^i$, $k = 1, 2, \ldots, K$ using majority vote i.e. $O^i = majority\{O_1^i, O_2^i, \ldots, O_K^i\}$, where $O^i$ is the predicted label of $i^{th}$ channel $X_i$.

Using the predictions of all channels, the second level ensemble predict the final state of the EEG signal $X$ using majority vote fusion, i.e. $O = majority\{O^1, O^2, \ldots, O^c\}$, where $O$ is the class label of the EEG Signal $X$.

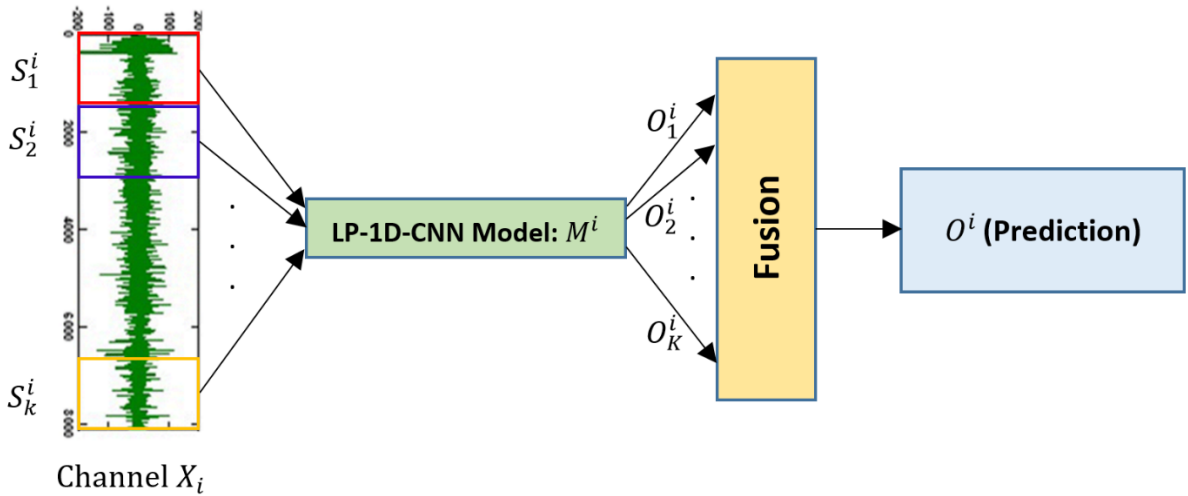

**FIGURE 1.** First level ensemble: non-overlapping windows of one EEG channel $X_i$ are passed to LP-1D-CNN model $M^i$ and their decisions are fused

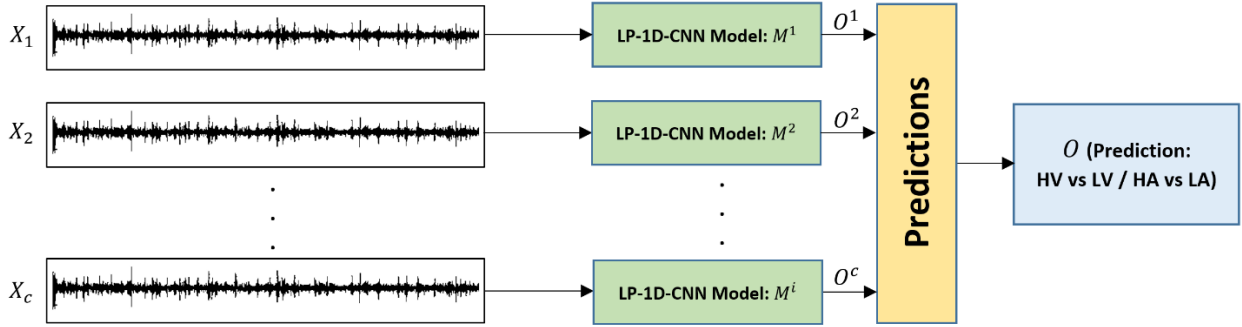

**FIGURE 2.** Proposed Architecture of Emotion Recognition System at second level ensemble

A brain signal evoked by a particular task is originated from a particular location of the brain. However, due to volume conduction, it is superimposed with other signals and is captured from different brain locations [44]. It means that different channels capture the signal evoked by a particular emotion but in different quantities. In view of this, we train a different CNN model for each channel so that it learns the part of the brain activity evoked by a particular emotion and is captured by the channel. Our hypothesis is that the fusion of the predictions of CNN models corresponding to different channels i.e. the second level ensemble will predict the emotion state represented by the EEG signal. Further, if we take the complete channel corresponding to an emotion state, it raises certain issues: (i) the available data is small and is not enough for a CNN model, (ii) the length of each channel is usually long e.g. in DEAP, the length of each channel is 8064. If the whole channel is used as input, it will cause to increase the depth and consequently, the number of learnable parameters of the CNN model and the overfitting will be unavoidable. To overcome these difficulties, we segment a channel into sub-signals using a window of fixed size, using these signals, we train a CNN model for the channel. It solves the above two problems. At test time, the CNN model will locally analyze a channel and fusion will give the global decision about the state of the channel. With these considerations, the designed LP-ID-CNN model for each channel has very low complexity i.e. only 8462 learnable parameters, and also, the data generated using windowing each channel is enough for its training and to avoid overfitting. In the following sections, we will discuss our proposed LP-1D-CNN model, data augmentation, training and testing schemes.

## 3.1  LP-1D-CNN Model

We used 1D-CNN for developing the Deep-AER system. The proposed LP-ID-CNN model is shown in Fig. 3; it consists of an input layer, convolutional (CONV) blocks and fully connected (FC) layers. The input layer takes a 1D channel of an EEG signal as input and passes it to a series of CONV blocks, which extract a hierarchy of features from the input signal. These features are passed to first FC layer, which further processes these features to extract the discriminative information and the finally the second FC layer together with softmax layer predict the class label of the input signal. The z-score normalization is used to normalize the input signals to unit variance and zero mean. This normalization helps in avoiding local minima and faster convergence. The normalized input is processed by four convolutional blocks, where each block consists of three layers: $Conv$ layer, batch normalization layer ($bN$) and non-linear activation layer ($Relu$). The number of kernels for $Conv$-1 is 32 and receptive field of each kernel is 1x5; the number of kernels for $Conv$-2 is 24 and receptive field and depth of each kernel is 1x3 and 32, respectively; the number of kernels for $Conv$-3 is 16 and receptive field and depth of each kernel is 1x3 and is 24, respectively; the number of kernels for $Conv$-4 is 8 and receptive field and depth of each kernel is 1x3 and is 16, respectively. The unnecessary or redundant features are reduced by using the bigger strides in ($Conv$) layers; the strides are 3, 2, 2 and 2 in $Conv$-1, $Conv$-2, $Conv$-3, and $Conv$-4, respectively. The output of the fourth block is passed to the first FC layer ($Fc1$) that is followed by a $Relu$ layer and another FC layer ($Fc2$). We examined two choices for the number of neurons in $Fc1$: 20 and 40. Dropout is used before $Fc2$ to avoid the risk of overfitting. Then, the output of $Fc2$ is passed to a softmax layer, which serves as a classifier and predicts the class of the input signal. Based on depending on the number of classes, the number of neurons in $Fc2$ is two (02). The specifications of this model and its variants are given in Table 1.

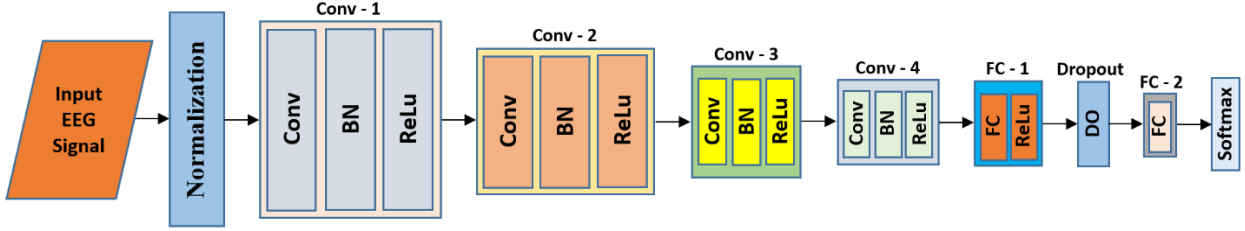

**FIGURE 3.** Proposed Architecture of Lightweight Pyramidal 1D-CNN (LP-1D-CNN) Model

**Table 1.** Specifications of four LP-1D-CNN models with Conv($1 \times r$, $/str, nic, noc$) [$1 \times r$ is receptive field, $str$ means stride, $nic$ and $noc$ are, respectively number of input and output feature maps (channels)]

| Layer | $M_1^i$ | $M_2^i$ | $M_3^i$ | $M_4^i$ |
|---|---|---|---|---|
| **Conv - 1** | (1×5, /3, 1, 32) | | | |
| | batchNorm | | | |
| | Relu | | | |
| **Conv - 2** | (1×3, /2, 32, 24) | | | |
| | batchNorm | | | |
| | Relu | | | |
| **Conv - 3** | (1×3, /2, 24, 16) | | | |
| | batchNorm | | | |
| | Relu | | | |
| **Conv - 4** | (1×5, /3, 16, 8) | | | |
| | batchNorm | | | |
| | Relu | | | |
| **FC1** | $Fc1 = 20$ | | $Fc1 = 40$ | |
| | Relu | | | |
| **Dropout** | - | 0.5 | - | 0.5 |
| **FC2** | $Fc2 = 2$ | | | |
| **Classifier** | Softmax | | | |
| **No. of Parameters** | 8462 | | 12842 | |

The proposed deep LP-ID-CNN model learns structures of EEG signals from data automatically and performs classification in an end-to-end manner. The proposed approach is opposite to the traditional hand-engineered approach, where first features are extracted, a subset of extracted features are selected and finally passed to a classifier for classification. The convolutional layer is the main component of a CNN model, which consists of a plane of many 1D channels or feature maps. In this layer, the convolution is performed by sliding the kernel over the input to obtain a convolved output (feature map). Let $a_{l-1} \in R^{N_{l-1} \times d_{l-1}}$ be the activation of the $(l-1)^{th}$ layer, where $d_{l-1}$ is the number of channels (feature maps) in the $(l-1)^{th}$ layer and $N_{l-1}$ is the number of neurons in each channel. Also, let $w_l^i \in R^{r_l \times d_{l-1}}$ be the $i^{th}$ kernel of $l^{th}$ layer, where $r_l$ is the receptive field of the kernel, then pre-activation of the $i^{th}$ channel of $l^{th}$ $Conv$ layer is calculated as follows:

$$c_l^i = w_l^i *_s a_{l-1} + b_l^i, \qquad (2)$$

where $b_l^i$ is the bias of the kernel $w_l^i$ and $*_s$ is the convolution operation with stride $s$. The activation of the channel is computed using ReLU non-linear activation function as follows:

$$a_l^i = ReLU(c_l^i). \qquad (3)$$

Please note that for the first $Conv$ layer, the input is $S_k^i = a_0 \in R^{N_0 \times 1}$, where $N_0$ is the number of sample points in the input 1D signal $S_k^i$. In the Fig. 3, the operation defined by the equation (2) is represented as $Conv$ layer and the operation defined by the equation (3) is represented as ReLU layer.

After $Conv$ blocks, each model $M^i$ has two $FC$ layers. All the neurons in $Conv4$ layer are connected to each neuron in the first fully connected layer $FC1$. The activation of $Conv4$ is $a_4 \in R^{N_4 \times d_4}$ or $a_4 \in R^{N_4 d_4}$ after vectorization. Let $W_1 \in R^{N_4 d_4 \times N_5}$ be the weight matrix of $FC1$ and $b_1 \in R^{N_5}$ be the bias vector of $FC1$, then its pre-activation is computed as follows:

$$z_1 = W_1^T a_4 + b_1 \qquad (4)$$

and its activation after applying ReLU non-linearity is calculated as follows:

$$a_5 = ReLU(z_1). \qquad (5)$$

Similarly the activation of $FC2$ is calculated. Note that the operations defined in equation (4) and (5) are represented as $FC1$ and ReLU layers in Fig. 3.

In different models, the number of neurons in $FC1$ are different, the detail is given in Table 1. The second fully connected layer $FC2$ has two neurons because HV vs LV or HA vs LA is a two class problem. Furthermore, outputs from the last fully connected layer are fed into softmax function to predict the class probability of the input EEG channel $X_i$. The further details regarding $Conv$ layers, batch normalization, ReLU and $FC$ layers can be found in [27, 45]. The 1D-CNN model analyzes a signal to learn a hierarchy of discriminative information and predict its class. In CNN, the kernels are learned from data unlike hand-engineered approach, where kernels are predefined, e.g., wavelet transform. CNN with its novel idea of shared kernels has the advantage of a significant reduction in the number of parameters.

Normally, a CNN model has small number of kernels in low-level layers and large number of kernels in high-level layers. However, the complexity of this type of structure is high due to large number of learnable parameters. The size of weight matrix $W_1$ in equation (4) depends on the number of neurons in the layer before the FC1 layer; if the neurons in the $Conv4$ block is large, then the size of $W_1$ is big i.e. it will cause a drastic increase in the number of learnable weights and it will lead to the problem of overfitting. Instead, we used a pyramid architecture, where number of kernels are large in low-level layers and small in higher-level layers. This architecture helps in avoiding the risk of overfitting by reducing the number of learnable parameters significantly. A large number of kernels are taken in a $Conv1$ layer, which are reduced by a constant number in $Conv2$, $Conv3$ and $Conv4$ layers, e.g., models $M^1$ to $M^4$, specified in Table 1, contain $Conv1$, $Conv2$, $Conv3$ and $Conv4$ layers with 32, 24, 16, and 8 kernels, respectively. The idea is that low-level layers extract a large number of microstructures, which are composed by higher level layers into higher level features which is small in number but discriminative i.e. it implicitly does the feature reduction and selection, which is an essential part of most of the methods based on hand-engineered features. In this study, four models are considered based on pyramid architecture to show the effectiveness of the LP-1D-CNN model. Table 1 shows detailed specifications of these models and also gives the number of learnable parameters in each model. With the help of these models, we show how a properly designed model can result in a better performance despite less parameters, which has less risk of overfitting. The models having pyramid architecture involve significantly less number of learnable parameters, see Table 1.

### 3.2   Data Augmentation

In our approach, the problem of predicting the state of an EEG signal $X$ is decomposed in smaller problems of predicting the classes of the cannels $X_i$ of the signal. If the whole channel $X_i$ is used as an input instance to a CNN model, then it is difficult to train the CNN model due to two reasons: (i) the complexity of the model is very high because of the input size (e.g. the length of each $X_i$ in DAEP is 8064 sample point) and (ii) the available data is not enough for training. To overcome this problem, $X_i$ is we divided into segments $S_k^i$ of temporal length $T_w$, which are passed to a CNN model to predict their states and fusing their sates, the class of $X_i$ is predicted. We use a window of $T_w$ seconds to create segments $S_k^i$. In this way, we need to train only one small CNN model for each $X_i$. The training instances of $X_i$ are segmented using a window of $T_w$ seconds to create training instances to train the corresponding LP-1D-CNN model $M^i$. In this way, we get enough training instances to train the model. The available instances of $X_i$ are divided into disjoint training and testing sets, which consist of 90% and 10% of total signals, respectively; only the training set is used to create training data for $M^i$.

In this study, we tested three different sizes for $T_w$: 5, 10 and 15 seconds. Using these sizes for $T_w$, we divided each channel $X_i$ of length 8064 samples into 12 sub-signals $S_k^i$ (672 samples each), 6 (1344 sample each) and 4 (2016 samples each), respectively. Out of the three choices, $T_w = 5$ seconds gives the best result for HV vs LV case over $FRONT$ brain region, i.e., 98.43%; however, using 10 and 15 seconds window sizes, the system gives accuracies of 96.8% and 93.7%, respectively. This shows that each small window of 5 seconds in the ensemble contains more relevant information and system analyses a local part of the signal minutely. Therefore, in all other experiments using different brain regions, i.e., *CENT, PERI, OCCIP* and *ALL*, we used window size of 05 seconds.

In the DEAP dataset, the total number of EEG signal instances is 1280 from 32 subjects. We divide these instances into training and testing sets so that 90% (i.e. 1152 EEG signals) is used for training and 10% (i.e. 128 EEG signals) for testing so that we can use 10-fold cross validation for performance evaluation. As the number of each channel $X_i$ in training EEG signals is 1152 and its length is 8064 samples, so the total number of instances of sub-signals $S_k^i$ corresponding to $X_i$ is 13824 if $T_w = 5$ i.e. 13824 patterns are available to train the LP-ID-CNN model $M^i$, which are enough to train it because it involves 8462 learnable parameters, see Table 1.

### 3.3 Training of LP-1D-CNN Model

Each LP-ID-CNN model $M^i$ is trained using the training data creased for the channel $X_i$, the detail is given in the previous section. For training the model, we used cross entropy loss function, stochastic gradient descent with Adam (SGDA) optimizer [46] and back-propagation for gradient calculation. According to SGDA, the learnable parameter, $\theta = (W, b)$ are updated using the following iterative procedure:

$$\theta_t = \theta_{t-1} - \alpha \frac{\hat{m}_t}{\sqrt{\hat{v}_t} + \varepsilon}, \tag{6}$$

where

$$\hat{m}_t = \frac{m_t}{1 - \beta_1^t}, \tag{7}$$

$$\hat{v}_t = \frac{v_t}{1 - \beta_2^t}, \tag{8}$$

$$m_t = \beta_1 \cdot m_{t-1} + (1 - \beta_1) \cdot g_t, \tag{9}$$

$$v_t = \beta_2 \cdot v_{t-1} + (1 - \beta_2) \cdot g_t^2, \tag{10}$$

and *m, v, t* and $g_t$ are the 1st moment vector, 2nd moment vector, timestep, and gradient of the loss function, respectively. This algorithm has four hyper-parameters: learning rate $(\alpha), \beta_1, \beta_2$ and epsilon $(\varepsilon)$. The parameters $\beta_1$ and $\beta_2$ represent the exponential decay rates. Following the recommendation of Kingma and Jimmy [46], in our experiments, we set $\beta_1 = 0.9$, $\beta_2 = 0.999$, $\alpha = 1 \times 10^{-4}$ and $\varepsilon = 10^{-8}$. It was observed that it enables the network to converge at a fast rate thereby improving the efficiency of the training process.

To improve the generalization and avoid overfitting, the dropout technique is applied to the $FC1$. In dropout, a probability value of 0.5 is used.

### 3.4 Testing

After training the model $M^i$ corresponding to each channel $X_i$, an unknown or test EEG signal $X$ is classified using two level ensemble of deep LP-ID-CNN models, the architectures of first and second level ensembles are shown in Fig. 1 and 2. The first level ensemble is designed to take decision about the state of each channel $X_i$. First the trained LP-1D-CNN model $M^i$ is used to predict the label of the channel $X_i$ by classifying each of its sub-signal $S_k^i$ and fusing their predictions. The temporal length of the channel $X_i$ in DEAP is 1 minute, which consists of 8064 samples; it is divided into $K$ non-overlapping sub-signals $S_k^i$ using a window of fixed temporal length $T_w$, which was used to create training patterns for learning $M^i$. These sub-signals $S_k^i$ are treated as independent signal instances and passed to LP-ID-CNN model $M^i$, predicts it class label i.e. $O_k^i = M^i(S_k^i)$ where $O_k^i \in \{0, 1\}$ is the predicted label of $S_k^i$, $k = 1, 2, \ldots, K$. The class label of $X_i$ is predicted by fusing the predictions of all sub-signals $S_k^i$ using majority vote i.e. $O^i = majority\{O_1^i, O_2^i, \ldots, O_K^i\}$, where $O^i$ is the predicted label of the channel $X_i$. Using the predictions of all channels obtained at first level ensemble, the second level ensemble predict the final state of the EEG signal $X$ using majority vote fusion, i.e. $O = majority\{O^1, O^2, \ldots, O^c\}$, where $O$ is the class label of the EEG Signal $X$.

## 4. Experimental Protocol and Evaluation Criteria

In this section, we present the detail of dataset used for experiments, experimental protocol and evaluation criteria.

### 4.1 Dataset

DEAP is a benchmark EEG database for the analysis of spontaneous emotions. This database was prepared by Queen Mary University of London [10]. It was created with the goal of creating an adaptive music video recommendation system based on user current emotion. This database was recorded by using music clips to evoke emotions in the participants. The database consists of physiological signals of 32 participants (16 men and 16 women, aged between 19 and 37, mean age: 26.9 years) recorded while watching 40 one-minute long music videos. The dataset contains 32 channel EEG signals and 8 peripheral physiological signals. The detail of EEG signal recordings, pre-processing and stimulus material can be found in [10]. At the end of each music clip, participants assessed their emotional states in terms of valence, arousal, dominance and liking. Self-Assessment Manikin (SAM) [47] was used to visualize for valence and arousal scales. Participants rated valence and arousal on a continuous 9-point scale. The valence scale is ranging from unhappy or sad to happy or joyful. The participants whose valence ratings were smaller than 5 were assumed to have negative emotions, while others were considered to have positive emotions. Similarly, the arousal scale is ranging from inactive or passive to active. The participants whose arousal ratings were smaller than 5 were considered as inactive whereas others were assumed as active. Sample EEG signals related to HV and LV state measured from Fp1 channel are shown in Fig. 4.

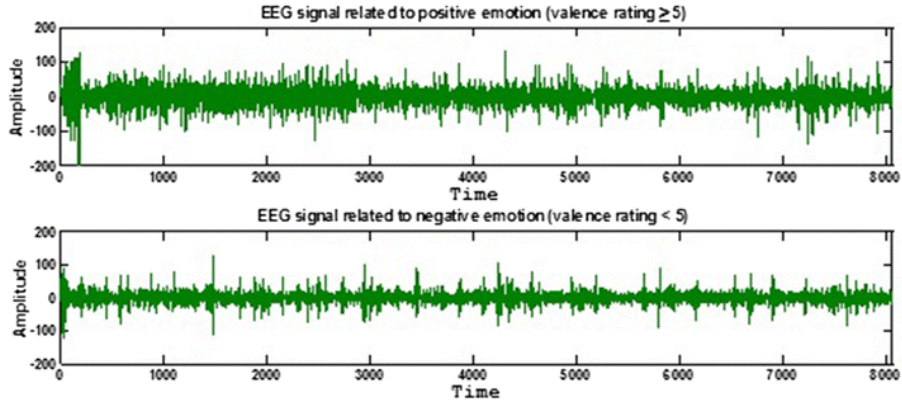

**FIGURE 4.** Sample EEG signals related to HV and LV valence. Sample EEG signals were captured from channel Fp1

In this study, we considered two dimensions, i.e., valence and arousal, and 32 channel EEG signals and addressed the problem of emotion recognition as two classification problems: HV vs LV and HA vs LA. The valence scale from 1 to 5 (excluding 5) is mapped to LV and 5 to 9 is mapped to HV. Similarly, the arousal scale from 1 to 5 (excluding 5) is mapped to LA and 5 to 9 is mapped to HA. In addition, we considered the following regions, as shown in Fig. 5, to identify their role in emotion recognition:

  i. $FRONT$: Frontal-right (FR) and frontal-left (FL) with 12 channels
 ii. $CENT$: Central-right (CR) and central-left (CL) with 4 channels
iii. $PERI$: Parietal-right (PR) and parietal-left (PL) with 6 channels
 iv. $OCCIP$: $O$ccipital − right (OR) and occipital-left (OL) with 4 channels
  v. *ALL:* All regions (AR) - 32 channels

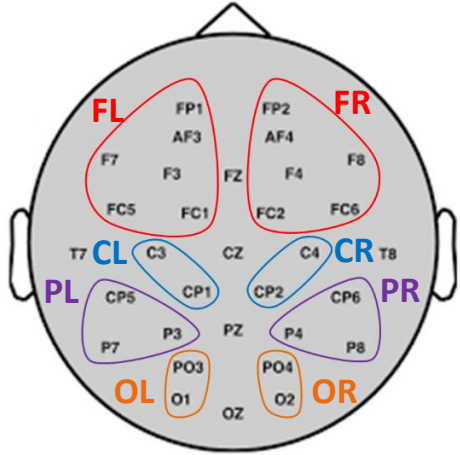

**FIGURE 5. Brain regions used for the analysis**. $FRONT$: frontal-right (FR) and frontal-left (FL); $CENT$: central-right (CR) and central-left (CL); $PERI$: parietal-right (PR) and parietal-left (PL); $OCCIP$: occipital-right (OR) and occipital-left (OL); $ALL$: All regions - 32 channels.

### 4.2 Evaluation Criteria

In order to evaluate the performance of the proposed system, 10-fold cross validation technique has been used to test the system over different variations of data. The signals for each class divided into 10 folds, each fold (10%), in turn, is kept for testing while the remaining 9 folds (90% signals) are used for learning the model. The average performance is computed for 10-folds. The well-known performance metrics were used to evaluate the performance such as accuracy, sensitivity, specificity, g-mean, f-measure, and precision. The definitions of these metrics are given below:

$$Accuracy\ (Acc) = \frac{TP + TN}{Total\ Samples} \tag{11}$$

$$Specificity\ (Spec) = \frac{TN}{TN + FP} \tag{12}$$

$$Sensitivity\ (Sens) = \frac{TP}{FN + TP} \tag{13}$$

$$Precision\ (Prec) = \frac{TP}{TP + FP} \tag{14}$$

$$F - Measure\ (FM)) = \frac{2 * Precision * Sensitivity}{Precision + Sensitivity} \tag{15}$$

$$G - Mean\ (GM) = \sqrt{Specificity * Sensitivity} \tag{16}$$

where *TP:* true positives is the number of LV/LA that are identified as LV/LA, *FN:* false negatives is the number of LV/LA that are predicted as HV/HA, *TN:* true negatives is the number of HV/HA that are identified as HV/HA by the system and *FP:* false positives is the number of HV/HA that are predicted as LV/LA.

TensorFlow was used to implement LP-ID-CNN model with python [48], a freely available deep learning library from Google. We trained LP-1D-CNN models $M^i$ on a desktop system with Intel (R) Xeon (R) CPU - E5-2670 v2 @ 2.5 GHz (20 CPUs) having 32GB RAM, 3GB Nvidia Quadro K4000 Graphics Card.

### 5. Experimental Results and Discussion

In this section, we present the results of two emotion classification problems i.e. HV vs LV and HA vs LA, and discuss them. We analyzed the potential of five different brain regions i.e. *FRONT, CENT, PERI, OCCIP* and *ALL* for emotion recognition using the proposed Deep-AER system. The best LP-1D-CNN model is selected by analyzing the results on different brain regions. The comparison is carried out with the state-of-the-art studies. The 10-fold cross validation technique has been used to perform all the experiments.

## 5.1 Specifications of LP-1D-CNN Models

To find the best model $M^i$ for each channel $X_i$, we considered four models $M^i_j$, $j = 1, 2, 3, 4$ as is shown in Table 1, and performed experiments for the five brain regions. We performed all the experiments using 10-fold cross validation with all the four models for the two problems: HV vs LV and HA vs LA. These experiments led us to select the best LP-1D-CNN model, which we used for onward analysis. DEAP dataset was used to train and test the models. Models $M^i_1$ to $M^i_4$ (pyramid models) are designed by reducing the number of kernels or filters $w^i_l$ by the ratio of 25% as the network goes deeper. The pyramid models contain less number of learnable parameters than traditional models, and as such are generalize well and less prone to overfitting.

## 5.2 Analysis of Brain Regions

To analyze which brain region is the most effective in emotion recognition, we performed 5 experiments considering the EEG signals captured from four different brain regions and the whole brain.

### 5.2.1 Experiment-1: *FRONT* Region

There are 12 channels $X_i$ in the EEG signal $X$ recorded from this region as shown in Fig. 5. The average performance results obtained with four different models ($M^i_1$ to $M^i_4$) on *FRONT* region are given in Table 2. It is observed that among the four models, when $M^i_2$ is used in Deep AER system, it yields the best mean accuracy of 98.43% for the problem of HV vs LV; the mean specificity and sensitivity are 97.8% and 98.7%, respectively. Similarly, for the problem of HA vs LA, the same model results in the best mean accuracy, which is 97.65%, whereas the mean specificity and sensitivity are 97.9% and 97.5%, respectively. The other performance measures for this model are better than those for other models.

The mean accuracies of the four models, when sub-signals $S^i_k$ of channel $X_i$, (i = 1,2,..., 12) of the FRONT region are used as training and testing instances, are shown in Fig. 6 and 9 for HV vs LV and HA vs LA. Further analysis of the mean training and testing accuracies of the four models on each channel of FRONT region (i.e. first level ensemble) and all channels of the region (i.e. second level ensemble) is shown in Fig. 7-8 for HV vs LV and in Fig 10-11 for HA vs LA. These results indicate that: (i) the model $M^i_2$ outperforms the other models in all cases; it is probably due to the reasons that it uses dropout and less number of neurons in *FC1*, which implicitly does the feature selection, (ii) all models do not suffer from over-fitting problem; the differences between mean training and testing accuracies are small, (iii) the ensemble enhances the performance of the model, when the decision is taken only using sub-signals $S^i_k$, the mean accuracy is less than that when first ensemble is used as is obvious from Fig. 6-7 and 9-10, further second level ensemble gives better mean accuracy than that by the first level ensemble as is clear from Fig. 7-8 and 9-10.

**Table 2.** Comparison of different LP-1D-CNN models over *FRONT* Region

|  | Model | $M^i_1$ | $M^i_2$ | $M^i_3$ | $M^i_4$ |
|---|---|---|---|---|---|
| HV vs LV | $Acc$ | 97.65±0.66 | **98.43±0.60** | 96.87±0.63 | 97.65±0.64 |
|  | $Sens$ | 0.975 | **0.987** | 0.975 | 0.975 |
|  | $Spec$ | 0.956 | **0.978** | 0.956 | 0.978 |
|  | $Prec$ | 0.975 | **0.988** | 0.975 | 0.987 |
|  | $GM$ | 0.965 | **0.982** | 0.965 | 0.976 |
|  | $FM$ | 0.975 | **0.987** | 0.975 | 0.981 |
| HA vs LA | $Acc$ | 96.87±0.68 | **97.65±0.63** | 96.1±0.66 | 96.87±0.70 |
|  | $Sens$ | 0.962 | **0.975** | 0.975 | 0.974 |
|  | $Spec$ | 0.959 | **0.979** | 0.94 | 0.959 |
|  | $Prec$ | 0.974 | **0.987** | 0.963 | 0.974 |
|  | $GM$ | 0.96 | **0.976** | 0.957 | 0.966 |
|  | $FM$ | 0.968 | **0.981** | 0.969 | 0.974 |

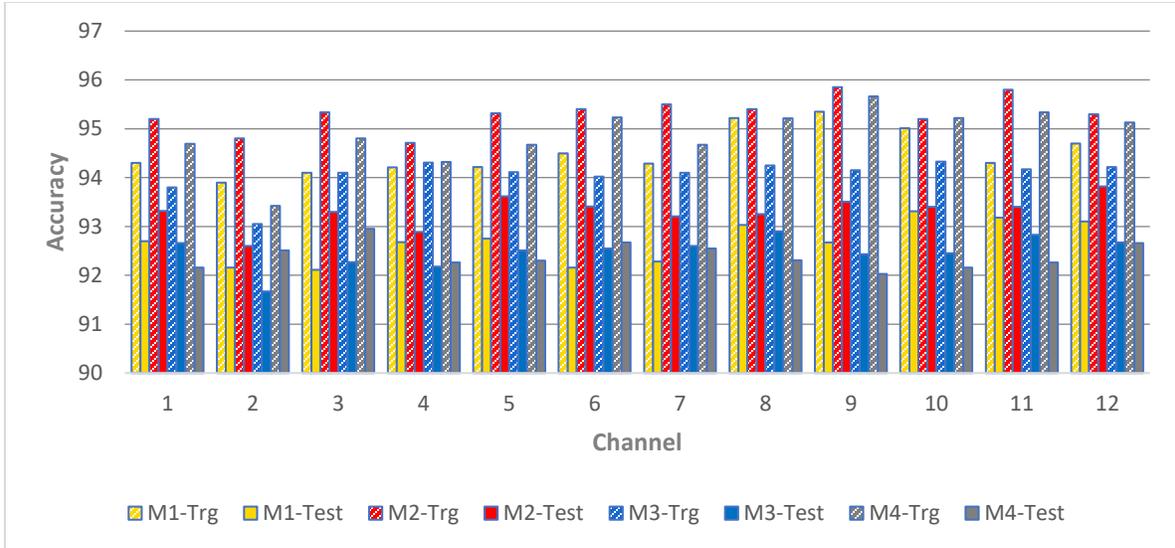

**FIGURE 6.** Single model channel-wise accuracies for HV vs LV with model $M_1^i$ to $M_4^i$ on the *FRONT* region.

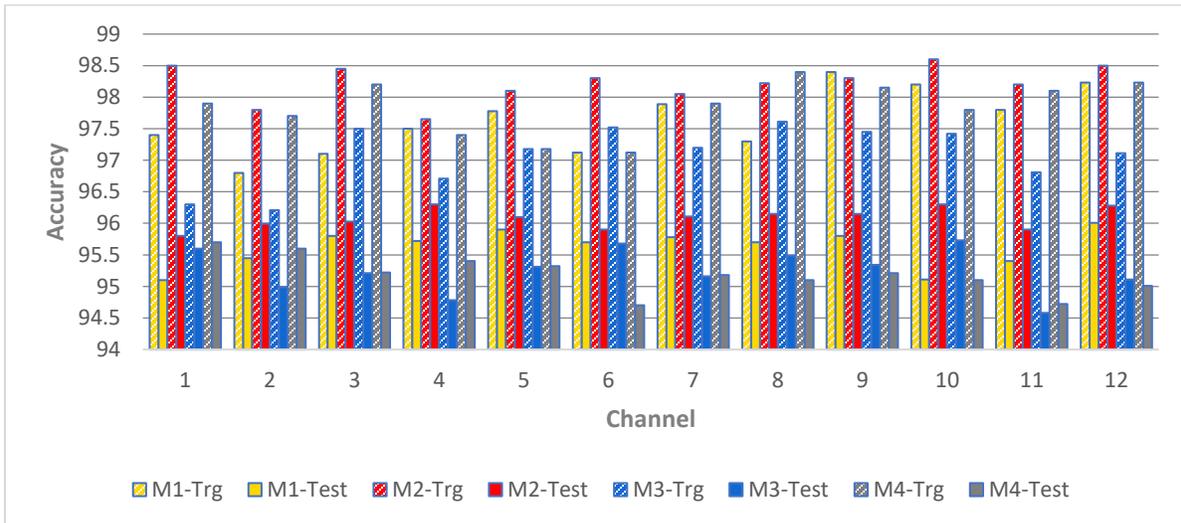

**FIGURE 7.** First ensemble level channel-wise accuracies for HV vs LV with models $M_1^i$ to $M_4^i$ on the *FRONT* region.

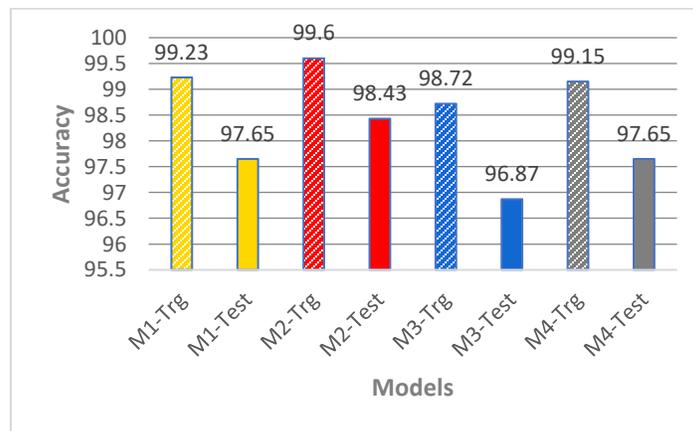

**FIGURE 8.** Second ensemble level accuracies of HV vs LV with models $M_1^i$ to $M_4^i$ on the *FRONT* region.

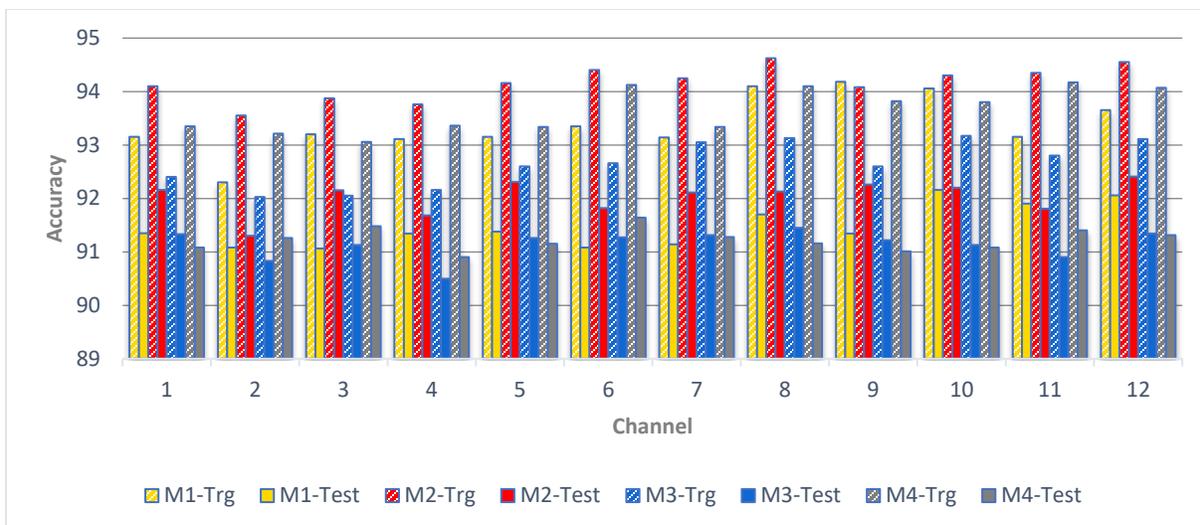

**FIGURE 9.** Single model channel-wise accuracies for HA vs LA with models $M_1^i$ to $M_4^i$ on the *FRONT* region.

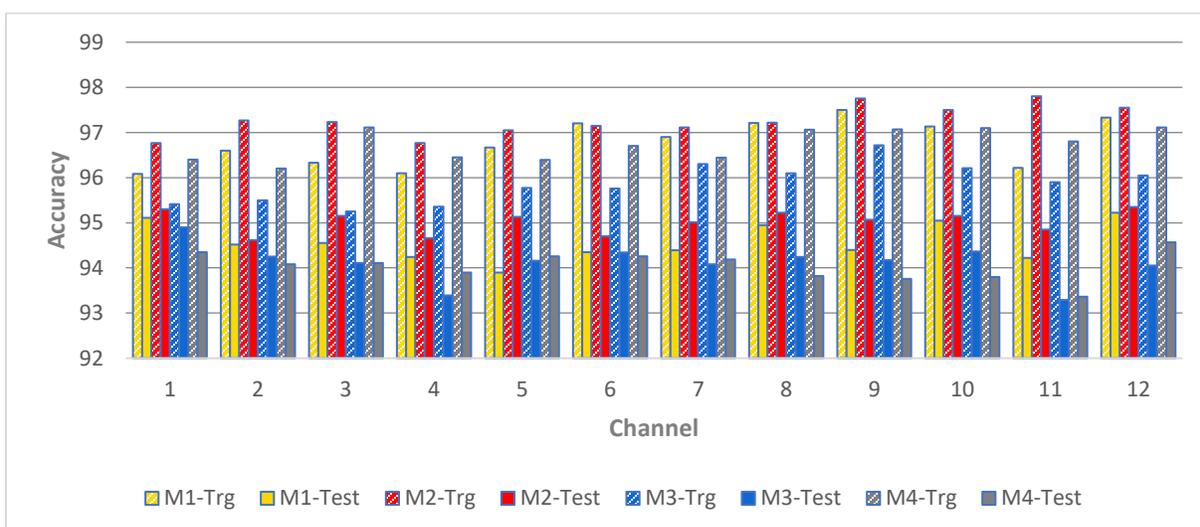

**FIGURE 10.** First ensemble level channel-wise accuracies for HA vs LA with models $M_1^i$ to $M_4^i$ on the *FRONT* region.

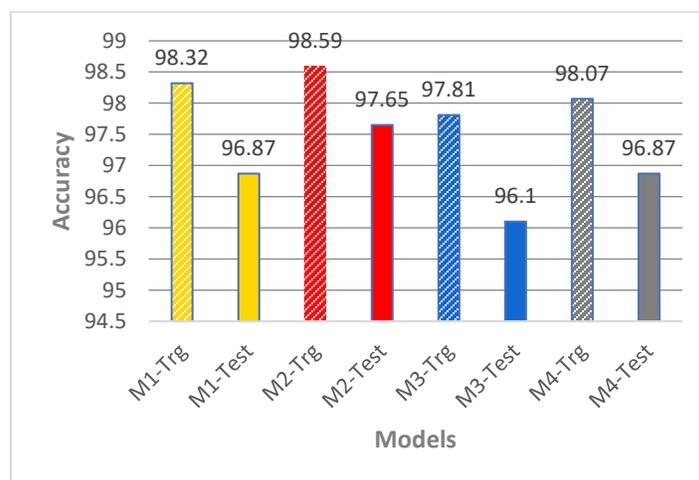

**FIGURE 11.** Second ensemble level accuracies for HV vs LV with models $M_1^i$ to $M_4^i$ on the *FRONT* region.

Based on overall results as shown in Table 2 and Fig. 6-11 and the above discussion, we conclude that the model $M_2^i$ with dropout layer and 20 neurons in the $FC1$ layer gives the best results when only the EEG signals from the *FRONT* region are considred. This model gives similar or slightly higher performance than other models but has less complexity i.e. it involves the less number of learnable parameters than the models $M_3^i$ and $M_4^i$. Further, to give insight into the performance of this model, the 10-fold cross-validation results on the *FRONT* region are shown in Table 3; the standard deviation is 0.60 for HV vs LV problem and 0.63 for HA vs LA problem, which gives indication of the robustness of the model. In view of the better performance of $M_2^i$, we will use only this model in all ownward experiments.

### 5.2.2 Experiment-2: *CENT* Region

The EEG signal $X$ recorded from this region has 4 channels as shown in Fig. 5. In this case, when $M_2^i$ is used in Deep AER system, it gives the mean accuracy of 92.3% (see Table 3) for the problem of HV vs LV; the mean specificity and sensitivity are 91.8% and 92.7%, respectively; the mean $Prec, GM$ and $FM$ are 92.7%, 91.9% and 93.2%, respectively. For the problem of HA vs LA, $M_2^i$ results in the mean accuracy of 93.8%, whereas the mean specificity and sensitivity are 94.5% and 93.2%, respectively; the mean $Prec, GM$ and $FM$ are 94.2%, 93.3% and 94.1%, respectively.

### 5.2.3 Experiment-3: *PERI* Region

There are 6 channels $X_i$ in the EEG signal $X$ recorded from this brain region as shown in Fig. 5. The accuracies obtained using $M_2^i$ on *PERI* region are shown in Table 3. It is observed that $M_2^i$ gives the mean accuracy of 94.6% for the problem of HV vs LV; the mean specificity and sensitivity are 93.8% and 95.7%, respectively. The mean $Prec, GM$ and $FM$ are 95.1%, 94.2% and 94.9%, respectively. In case of HA vs LA, $M_2^i$ results in the mean accuracy of 93.2%, whereas the mean specificity and sensitivity are 94.1% and 92.4%, respectively; the mean $Prec, GM$ and $FM$ are 93.7%, 92.6% and 93.5%, respectively.

### 5.2.4 Experiment-4: *OCCIP* Region

The EEG signal $X$ recorded from this region has 4 channels as shown in Fig. 5. The accuracies obtained using $M_2^i$ on this region are shown in Table 3; it gives the mean accuracy of 91.4% for the problem of HV vs LV; the mean specificity and sensitivity are 90.8% and 92.7%, respectively. The mean $Prec, GM$ and $FM$ are 91.8%, 90.6% and 91.7%, respectively. In case of HA vs LA problem, $M_2^i$ results in the mean accuracy of 92.7%, whereas the mean specificity and sensitivity are 93.4% and 92.1%, respectively; the mean $Prec, GM$ and $FM$ are 93.6%, 91.4% and 93.2%, respectively.

### 5.2.5 Experiment-5: *ALL* Regions

In this experiment, the EEG signal $X$ consists of 32 channels covering all brain regions as shown in Fig. 5. The accuracies obtained using $M_2^i$ are shown in Table 3. It is observed that when $M_2^i$ is used in Deep AER system, it gives the mean accuracy of 91.7% for the problem of HV vs LV; the mean specificity and sensitivity are 90.4% and 92.8%, respectively; the mean $Prec, GM$ and $FM$ are 92.6%, 91.1% and 92.2%, respectively. In case of HA vs LA, $M_2^i$ results in the mean accuracy of 90.3%, whereas the mean specificity and sensitivity are 91.8% and 90.1%, respectively; the mean $Prec, GM$ and $FM$ are 91.2%, 89.6% and 90.9%, respectively.

**Table 3**. The 10-fold cross validation accuracies (%) of the Deep-AER system on different brain regions using model $M_2^i$ for two-class problems

| | HV vs LV | | | | | HA vs LA | | | | |
|---|---|---|---|---|---|---|---|---|---|---|
| Fold | *FRONT* | *CENT* | *PERI* | *OCCIP* | *ALL* | *FRONT* | *CENT* | *PERI* | *OCCIP* | *ALL* |
| $K1$ | 98.43 | 92.9 | 95.3 | 91.4 | 89.8 | 96.8 | 94.5 | 93.7 | 93.7 | 91.4 |
| $K2$ | 99.2 | 91.4 | 93.7 | 92.1 | 93.7 | 98.43 | 92.9 | 92.9 | 91.4 | 90.6 |
| $K3$ | 98.43 | 92.1 | 95.3 | 91.4 | 92.1 | 97.65 | 93.7 | 93.7 | 93.7 | 89.8 |
| $K4$ | 97.65 | 92.9 | 94.5 | 90.6 | 89.8 | 97.65 | 94.5 | 92.9 | 92.1 | 90.6 |
| $K5$ | 98.43 | 92.1 | 95.3 | 92.1 | 92.1 | 98.43 | 93.7 | 93.7 | 91.4 | 89.8 |
| $K6$ | 99.2 | 93.7 | 94.5 | 91.4 | 91.4 | 96.8 | 94.5 | 92.1 | 93.7 | 90.6 |
| $K7$ | 97.65 | 90.6 | 93.7 | 92.1 | 90.6 | 97.65 | 93.7 | 93.7 | 91.4 | 89.1 |

| | | | | | | | | | |
|---|---|---|---|---|---|---|---|---|---|
| **K8** | 99.2 | 91.4 | 94.5 | 90.6 | 93.7 | 98.43 | 92.9 | 91.4 | 92.9 | 89.8 |
| **K9** | 98.43 | 93.7 | 95.3 | 92.1 | 92.1 | 97.65 | 94.5 | 94.5 | 93.7 | 90.6 |
| **K10** | 97.65 | 92.1 | 94.4 | 90.6 | 91.4 | 96.8 | 93.7 | 93.7 | 92.9 | 91.4 |
| **Mean** | **98.43** | 92.3 | 94.6 | 91.4 | 91.7 | **97.65** | 93.8 | 93.2 | 92.7 | 90.3 |
| **STD** | **0.60** | 0.96 | 0.60 | 0.62 | 1.30 | **0.63** | 0.59 | 0.86 | 0.97 | 0.70 |

To assess the performance of Deep-AER system, we conducted five exepriments corresponding to five diferent brain regions, as shown in Fig. 5. The number of channels in the EEG signals captured from *FRONT, CENT, PERI, OCCIP* and *ALL* are 12, 4, 6, 4 and 32, respectively. In each experiment, after training the LP-1D-CNN models $M_2^i$ for each channel $X_i$ of the corresponding region, we designed a Deep-AER system as a two-level ensemble, which employs majority vote strategy to fuse the local decisions at each level for HV vs LV and HA vs LA problems. The different regions lead to different results as shown in Tables 3; the Deep-AER system gives the best performance on *FRONT* region with model $M_2^i$; the accuracies on all other regions are below 95%; the comparison of different brain regions w.r.t accuracy is shown in Fig. 12. The results indicate that the *FRONT* region plays dominant role in emotion recognition; it gives the accuricies of 98.43% and 97.65% for HV vs LV and HA vs LA problems, respectively; the detailed results for the *FRONT* region with model $M_2^i$ are shown in Table 3 and Fig. 6-11. The 10-fold cross validation results shown in Table 3 point out that standard deviations in case of *FRONT* for the two problems are 0.60 and 0.63, whereas the standard deviations for other regions are higher except *CENT* for HA vs LA problem. It means that the *FRONT* results in a robust Deep-AER system for emotion recogntion i.e. the system gives almost similar results over variations of training and testing datasets.

The channel-wise and first level ensemble results depicted in Fig. 6-7 and 9-10 show that the ensemble performs better than the single model, the reason is that in the ensemble a model simulates experts analysing local parts of the signal, and finally their local decisions are fused using majority vote to take the final decision. In this way, the first level ensemble combines the local decisions with global context and outperforms a single model. Further, the Fig. 7-8 and 10-11 show that the second level enseble outperforms the first level ensemble; it is due the reason that second level ensemble fuses the local decisions based on individual channels with global context defined by all channels in a specific brain region. Also, the Deep-AER system is based on an end-to-end LP-1D-CNN models i.e. each model takes input signal and gives the decision; there is no need of any kind of signal preprocessing, manual feature extraction and selection and laborious parameter tuning. It learns the discriminative information automatically from the data and the learning process is fully automatic.

It is to be noted that our design of LP-1D-CNN model requires minimum memory space; the architecture of a LP-1D-CNN model is based on pyramid design, which involves the lowest number of learnable parameters. The best pyramid based LP-1D-CNN model $M_2^i$ contains 8462 parameters. The small number of learable parameters means less complex model, which not only results in less memory overhead, but also ensures better generalization. It implies that the proposed Deep-AER system does not heavily depend on data, is robust and has better generalization than state-of-the-art methods. The mean accuracy of the Deep-AER system on *FRONT* region with model $M_2^i$ is 98.43% and 97.65% for HV vs LV and HA vs LA, respectively, which validates the generalization power of the proposed system. Tables 4 and 5 show the confusion matrices for HV vs LV and HA vs LA problems on the *FRONT* region.

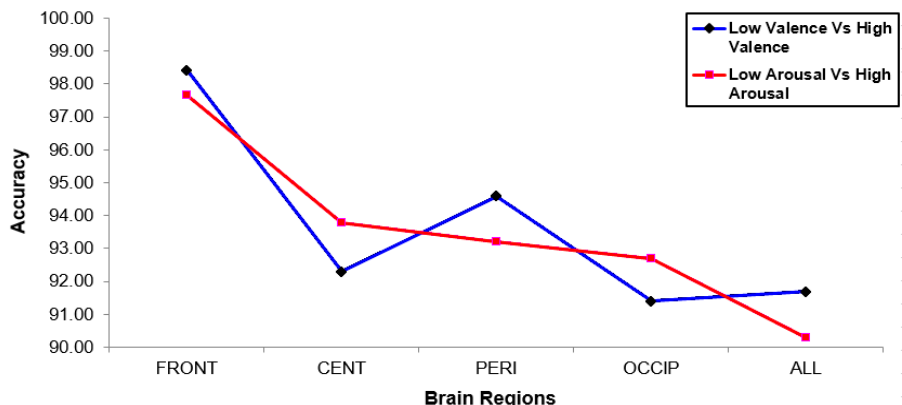

**FIGURE 12**. Comparison of different brain regions w.r.t accuracy

**Table 4.** Confusion matrix for HV vs LV problem on the *FRONT* region using model $M_2^i$

|  |  | Predicted Class | |
|---|---|---|---|
|  |  | **Low** | **High** |
| **Actual Class** | **Low** | 81 (98.78%) | 1 (1.22%) |
|  | **High** | 1 (2.17%) | 45 (97.83%) |

**Table 5.** Confusion matrix for HA vs LA problem on the *FRONT* region using model $M_2^i$

|  |  | Predicted Class | |
|---|---|---|---|
|  |  | **Low** | **High** |
| **Actual Class** | **Low** | 77 (97.47%) | 2 (2.53%) |
|  | **High** | 1 (2.04%) | 48 (97.96%) |

### 5.3  Analysis of Dominant Brain Region in Deep-AER System

In previous studies [49-53], researchers highlighted the importance and association of *FRONT* brain region with emotions. In the neuroscience research literature, this association has been extensively discussed. Based on the results obtained in this study, we also observed the dominance of *FRONT* region as compared to other brain regions, i.e., *CENT*, *PERI*, *OCCIP* and *ALL* for HV vs LV and HA vs LA classification problem. Our findings validates the previous research findings on the involvement of the *FRONT* region in positive and negative emotions [49-53] and we got the best accuracy rates on the *FRONT* region for HV vs LV and HA vs LA problems. All of the other brain regions, i.e., *CENT*, *PERI*, *OCCIP* and *ALL* give the accuracies less than 95% for HV vs LV and HA vs LA problems. The question arises that whether the channels of the EEG signal captured from the *FRONT* region are correlated or not. Table 6 shows the pearson correlation coefficients ($r$) between the channels on *FRONT* region for HV vs LV problem. It is observed that there are few correlation coefficients which show moderate positive or negative correlation. Most of the correlation coefficients are in the range of weak and negligible positive or negative correlation. There are no strong or very strong correlation coefficients between the channels over the *FRONT* region when using model $M_2^i$. Therefore, the maximum correlation between the channels found to be from negligible to weak correlation. It means that all the channels in *FRONT* region have discriminative information and must be considered for the design of Deep-AER system.

**Table 6.** Pearson Correlation Coefficients ($r$) between Channels for HV vs LV over *FRONT* Region using model $M_2^i$

|  | $C_1$ | $C_2$ | $C_3$ | $C_4$ | $C_5$ | $C_6$ | $C_7$ | $C_8$ | $C_9$ | $C_{10}$ | $C_{11}$ | $C_{12}$ |
|---|---|---|---|---|---|---|---|---|---|---|---|---|
| $C_1$ | 1.0000 | -0.0192 | -0.0562 | 0.1272 | 0.1877 | -0.0238 | -0.0780 | -0.1905 | -0.2614 | -0.3589 | -0.0185 | 0.1165 |
| $C_2$ | -0.0192 | 1.0000 | -0.2524 | -0.0187 | -0.0650 | 0.2223 | 0.0578 | 0.0877 | 0.1995 | 0.3217 | 0.0467 | 0.3609 |
| $C_3$ | -0.0562 | -0.2524 | 1.0000 | -0.1931 | -0.1407 | -0.0268 | 0.1307 | 0.0040 | 0.0990 | 0.1861 | -0.0825 | 0.2768 |
| $C_4$ | 0.1272 | -0.0187 | -0.1931 | 1.0000 | -0.1670 | -0.0973 | -0.1024 | 0.0452 | 0.1458 | -0.2279 | -0.1670 | -0.1383 |
| $C_5$ | 0.1877 | -0.0650 | -0.1407 | -0.1670 | 1.0000 | -0.0142 | 0.2765 | -0.1571 | -0.2156 | -0.0056 | 0.0032 | 0.2417 |
| $C_6$ | -0.0238 | 0.2223 | -0.0268 | -0.0973 | -0.0142 | 1.0000 | -0.1408 | -0.1079 | -0.0197 | 0.1154 | 0.0126 | 0.1887 |
| $C_7$ | -0.0780 | 0.0578 | 0.1307 | -0.1024 | 0.2765 | -0.1408 | 1.0000 | -0.2224 | -0.1141 | 0.0442 | 0.1300 | -0.3138 |
| $C_8$ | -0.1905 | 0.0877 | 0.0040 | 0.0452 | -0.1571 | -0.1079 | -0.2224 | 1.0000 | -0.0430 | 0.1044 | 0.0768 | 0.1995 |
| $C_9$ | -0.2614 | 0.1995 | 0.0990 | 0.1458 | -0.2156 | -0.0197 | -0.1141 | -0.0430 | 1.0000 | 0.0439 | -0.1042 | 0.2928 |
| $C_{10}$ | -0.3589 | 0.3217 | 0.1861 | -0.2279 | -0.0056 | 0.1154 | 0.0442 | 0.1044 | 0.0439 | 1.0000 | -0.1330 | 0.1814 |
| $C_{11}$ | -0.0185 | 0.0467 | -0.0825 | -0.1670 | 0.0032 | 0.0126 | 0.1300 | 0.0768 | -0.1042 | -0.1330 | 1.0000 | 0.2261 |
| $C_{12}$ | 0.1165 | 0.3609 | 0.2768 | -0.1383 | 0.2417 | 0.1887 | -0.3138 | 0.1995 | 0.2928 | 0.1814 | 0.2261 | 1.0000 |

## 5.4 Comparisons

To evaluate the effectiveness of the proposed Deep-AER system, we compared our experimental results with those of previous studies for emotion recognition based on EEG signals. The existing methods as shown in Table 7 use hand-engineered features such as PSD, power asymmetry, band power, statistical features, QTFD and EMD based features, fractal dimension, Hjorth parameters, wavelet statistical features, EEG spectral power and wavelet entropy for HV vs LV and HA vs LA problems. From Table 7, it is clear that the proposed Deep-AER system shows better classification performance as compared to other state-of-the-art approaches and the difference is significant. The Deep-AER system outperforms the existing methods due to three reasons. First, it is based on deep learning approach, which has shown outstanding performance for many problems as compared to hand-engineered features [17-22]. Second, it employs a pyramid architecture for the design of CNN models, which has less complexity and does not require big data for its learning. Third, it uses the ensemble strategy, which combines the local decisions with global context.

**Table 7.** Performance Comparison of the Proposed System for HV vs LV and HA vs LA

| Research | Features | Classifier | Accuracy HV vs LV | Accuracy HA vs LA |
|---|---|---|---|---|
| Koelstra et al. [10], 2012 | PSD | Gaussian Naive Bayes | 57.6% | 62.0% |
| Chung and Yoon [32], 2012 | PSD | Naive Bayes | 66.6 % | 66.4 % |
| Haung et al. [12], 2012 | Asymmetry spatial pattern (ASP) | KNN, Naive Bayes, SVM | 66.05% | 82.46% |
| Zhang et al. [36], 2013 | PSD, Statistical features | Ontological model | 75.19% | 81.74% |
| Rozgic et al. [34], 2013 | PSD | SVM | 76.9% | 69.1% |
| Candra et al. [33], 2015 | Wavelet entropy | SVM | 65.13% | 65.33% |
| Atkinson and Campos [38], 2016 | Band power, statistical, fractal dimension and Hjorth parameters features | SVM | 73.1% | 73.0% |
| Liu et al. [37], 2016 | Deep belief network (DBN) based features | SVM | 85.2% | 80.5% |
| Abeer et al. [35], 2017 | PSD, Frontal asymmetry | DNN | 82.0% | 82.0% |
| Tripathi et al. [39], 2017 | Statistical time domain features | Neural Networks (NN) | 81.4% | 73.3% |
| Zhuang et al. [41], 2017 | EMD based features | SVM | 69.1% | 71.9% |
| Li et al. [42], 2017 | Frequency-domain, non-linear dynamic domain and time-domain features | SVM | 80.7% | 83.7% |
| Yin et al. [40], 2017 | Power spectral and Statistical features | Neural Networks (NN) | 83.04% | 84.18% |
| Menezes et al. [43], 2017 | PSD, Higher order crossings (HOC) and Statistical features | SVM | 88.4% | 74.0% |
| Alazrai et al. [11], 2018 | QTFD-based features | SVM | 85.8% | 86.6% |
| **The proposed work** | Features extracted using LP-1D-CNN model | Softmax | 98.43% | 97.65% |

## 6. Conclusion

We addressed the problem of emotion recognition from EEG brain signals and modeled it as two binary classification problems i.e. HV vs LV and HA vs LA based on dimensional approach for emotion modeling [3]. For HV vs LV and HA vs LA problems, we developed a Deep-AER system based on deep LP-1D-CNN models and validated it using benchmark DEAP dataset. Most of the existing research studies on this problem are based on hand-engineered features that involve laborious manual parameter tuning, their performance heavily depends on the selection of hyper-parameters; they do not learn the internal structure of the data. As such, they do not generalize well across different cases. In addition, they involve laborious designs, i.e., first features are extracted and selected and then passed to a classifier, all these stages involve hyper-parameters whose joint manual tuning is laborious and time consuming. In contrast, we proposed a deep LP-1D-CNN model, which contains a small number of learnable parameters, which are learned in an end-to-end fashion; this model automatically and implicitly extract and select features, and finally classifies them. Using LP-1D-CNN, we build a two level ensemble model. In the first level of the ensemble, each channel is scanned incrementally by LP-1D-CNN to generate predictions, which are fused using majority vote. The second level of the ensemble combines the predictions of all channels of an EEG signal using majority vote for detecting the emotion state. To identify the brain region that has dominant role in AER, we analyzed EEG signals over five brain regions: *FRONT, CENT, PERI, OCCIP* and *ALL*. The results indicate that *FRONT* plays dominant role in AER and over this region, Deep-AER achieved the accuracies of 98.43% and 97.65% for two AER problems, i.e., HV vs LV and HA vs LA, respectively. The Deep-AER system has substantial improvements over prior systems for emotion recognition based on EEG signals as it outperforms the state-of-the-art techniques by large margin. More importantly, it shows that deep learning based system for the classification of brain signals outperforms traditional techniques. The results show that the deep learning based method provides better classification performance as compared to other state-of-the-art approaches and suggest that this method can be applied successfully to develop other EEG based expert systems. There are many future directions related to the proposed work. One of the future directions is to develop the system for the identification of individual emotions. Though the proposed system gives good performance on a benchmark dataset, its deployment in real-time environment for healthcare sector and security domains is also a future work. Further, the deep model can be extended to design a more generalized and powerful model by increasing the depth of the model.

## References

1. Liu, Yisi, and Olga Sourina. "EEG-based subject-dependent emotion recognition algorithm using fractal dimension." Systems, Man and Cybernetics (SMC), 2014 IEEE International Conference on IEEE, (2014).
2. Sreeshakthy, M., and J. Preethi. "Classification of Human Emotion from Deap EEG Signal Using Hybrid Improved Neural Networks with Cuckoo Search." BRAIN. Broad Research in Artificial Intelligence and Neuroscience6.3-4 (2016): 60-73
3. Grandjean, D., Sander, D., & Scherer, K. R. (2008). Conscious emotional experience emerges as a function of multilevel, appraisal-driven response synchronization. Consciousness and Cognition, 17(2), 484–495. doi:10.1016/j.concog.2008.03.019
4. Darwin, C. (1998). The expression of the emotions in man and animals (3rd ed.). New York: Oxford University Press.
5. Ekman, P. (1982). Emotion in the human face. Cambridge, UK: Cambridge University Press.
6. Davitz, J. (1964). Auditory correlates of vocal expression of emotional feeling. In J. Davitz (Ed.), The communication of emotional meaning (pp. 101-112). New York: McGraw-Hill.
7. Mehrabian, A., & Russell, J. (1974). An approach to environmental psychology. Cambridge, MA: MIT Press.
8. Osgood, C., Suci, G., & Tannenbaum, P. (1957). The measurement of meaning. Chicago: University of Illinois Press.
9. Hatice Gunes, Maja Pantic, "Automatic, Dimensional and Continuous Emotion Recognition", International Journal of Synthetic Emotions, Volume 1 Issue 1, Pages 68-99, January 2010.
10. S. Koelstra, C. Mühl, M. Soleymani, J. S. Lee, A. Yazdani, T. Ebrahimi, T. Pun, A. Nijholt, and I. Patras, "DEAP: A database for emotion analysis; Using physiological signals", IEEE Transactions on Affective Computing, vol. 3, no. 1, pp. 18–31, 2012.
11. Rami Alazrai, Rasha Homoud, Hisham Alwanni and Mohammad I. Daoud, "EEG-Based Emotion Recognition Using Quadratic Time-Frequency Distribution", Sensors 2018, 18, 2739; doi:10.3390/s18082739.
12. D. Huang, C. Guan, K. K. Ang, H. Zhang, and Y. Pan, "Asymmetric spatial pattern for EEG-based emotion detection," in Proceeding of the International Joint Conference on Neural Networks (IJCNN '12), pp. 1–7, Brisbane, Australia, June 2012.
13. G. Chanel, J. J.M. Kierkels, M. Soleymani, and T. Pun, "Shortterm emotion assessment in a recall paradigm," International Journal of Human Computer Studies, vol. 67, no. 8, pp. 607–627, 2009.
14. D. Nie, X.-W. Wang, L.-C. Shi, and B.-L. Lu, "EEG-based emotion recognition during watching movies," in Proceedings of the 5th International IEEE/EMBS Conference on Neural Engineering (NER '11), pp. 667–670, Cancun, Mexico, May 2011.
15. X.-W. Wang, D . Nie, and B.-L. Lu, "EEG-based emotion recognition using frequency domain features and support vector machines," in Neural Information Processing, B.-L. Lu, L. Zhang, and J. Kwok, Eds., vol. 7062, pp. 734–743, Springer, Berlin, Germany, 2011.
16. N. Jatupaiboon, S. Pan-ngum, and P. Israsena, "Real-time EEG based happiness detection system," The ScientificWorld Journal, vol. 2013, Article ID618649, 12 pages, 2013.

17. Eman Albilali, Hatim Aboalsamh, Areej Al-Wabil, "Comparing Brain-Computer Interaction and Eye Tracking as Input Modalities: An Exploratory Study". International Conference on Current Trends in Information Technology, Dubai (CTIT Dubai 2013), pp. 232-236, 11-12 Dec 2013.
18. Yann. LeCun and Yoshua. Bengio. Convolutional networks for images, speech, and timeseries. In M. A. Arbib, editor, The Handbook of Brain Theory and Neural Networks. MIT Press, 1995.
19. Alex Krizhevsky and Sutskever, Ilya and Hinton, Geoffrey E, Imagenet Classification with Deep Convolutional Neural Networks, in NIPS 2012, pp: 1097-1105
20. Simonyan, K., Zisserman, A.: Very Deep Convolutional Networks for Large-Scale Image Recognition. Corr abs/1409.1556 (2014)
21. Ji, S., Yang, M., Yu, K.: 3D Convolutional Neural Networks for Human Action Recognition. IEEE Transaction on Pattern Analysis and Machine Intelligence 35(1) (2013) 221{31}
22. Tran, D., Bourdev, L., Fergus, R., Torresani, L., Paluri, M.: Learning Spatiotemporal Features with 3D Convolutional Networks. In: ICCV, IEEE (2015) 1725{1732
23. Lecun, Y., Bottou, L., Bengio, Y., & Haffner, P. (1998). Gradient-based learning applied to document recognition. Proceedings of the IEEE, 86(11), 2278–2324. Http://doi.org/10.1109/5.726791
24. Cui, Z., & Chen, Y. (n.d.). Multi-Scale Convolutional Neural Networks for Time Series Classification.
25. Zhang, X., & Lecun, Y. (2015). Text Understanding from Scratch.
26. T. Ince, S. Kiranyaz, L. Eren, M. Askar and M. Gabbouj, Real-Time Motor Fault Detection by 1-D Convolutional Neural Networks, in IEEE Transactions on Industrial Electronics, vol. 63, no. 11, pp. 7067-7075, Nov. 2016.
27. Ihsan Ullah, Muhammad Hussain, Emad-ul-Haq Qazi, Hatim Aboalsamh, "An automated system for epilepsy detection using EEG brain signals based on deep learning approach", Expert Systems With Applications 107 (2018) 61–71.
28. T. Zhang, W. Chen, and M. Li, "AR based quadratic feature extraction in the VMD domain for the automated seizure detection of EEG using random forest classifier", in Biomedical signal processing and control, 2016
29. Sharmila, A., & Geethanjali, P, "DWT Based Detection of Epileptic Seizure From EEG Signals Using Naive Bayes and k-NN Classifiers", Vol. 4, 2016
30. Andrzejak, R. G., Lehnertz, K., Mormann, F., Rieke, C., David, P., & Elger, C. E. (2001). Indications of nonlinear deterministic and finite-dimensional structures in time series of brain electrical activity : Dependence on recording region and brain state, 64, 1–8. Http://doi.org/10.1103/physreve.64.061907
31. E. Qazi, M. Hussain, H. Aboalsamh, W. Abdul, S. Bamatraf, I. Ullah, "An Intelligent System to Classify Epileptic and Non-Epileptic EEG Signals", in Proc. SITIS, 2016
32. Chung, Seong Youb, and Hyun Joong Yoon. "Affective classification using Bayesian classifier and supervised learning." Control, Automation and Systems (ICCAS), 2012 12th International Conference on. IEEE, (2012).
33. Henry Candra, Mitchell Yuwono, Rifai Chai, Ardi Handojoseno, Irraivan Elamvazuthi, Hung T. Nguyen, and Steven Su, "Investigation of window size in classification of EEG-emotion signal with wavelet entropy and support vector machine." Engineering in Medicine and Biology Society (EMBC), 2015 37th Annual International Conference of IEEE, (2015).
34. Rozgic, Viktor, Shiv N. Vitaladevuni, and Ranga Prasad. "Robust EEG emotion classification using segment level decision fusion." Acoustics, Speech and Signal Processing (ICASSP), 2013 IEEE International Conference on IEEE, (2013).
35. Abeer Al-Nafjan, Areej Al-Wabil, Manar Hosny, Yousef Al-Ohali, "Classification of Human Emotions from Electroencephalogram (EEG) Signal using Deep Neural Network", IJACSA) International Journal of Advanced Computer Science and Applications, Vol. 8, No. 9, 2017
36. X. Zhang, Hu, J. hen, and Philip Moore, "Ontology-based context modeling for emotion recognition in an intelligent web", World Wide Web, vol. 16, no. 4, pp. 497–513, 2013.
37. Liu, W., Zheng, W.L., Lu, B.L.: Emotion recognition using multimodal deep learning. In: Hirose, A., Ozawa, S., Doya, K., Ikeda, K., Lee, M., Liu, D. (eds.) ICONIP 2016. LNCS, vol. 9948, pp. 521–529. Springer, Cham (2016). doi:10.1007/978-3-319-46672-9 58
38. J. Atkinson and D. Campos, ―Improving I-based emotion recognition by combining EEG feature selection and kernel classifiers. Expert Systems with Applications, vol. 47, pp. 35–41, 2016.
39. Samarth Tripathi, Shrinivas Acharya, Ranti Dev Sharma, Sudhanshi Mittal, Samit Bhattacharya, "Using Deep and Convolutional Neural Networks for Accurate Emotion Classification on DEAP Dataset", Proceedings of the Twenty-Ninth AAAI Conference on Innovative Applications (IAAI-17))
40. Yin, Z., Zhao, M., Wang, Y., Yang, J., Zhang, J.: Recognition of emotions using multimodal physiological signals and an ensemble deep learning model. Comput. Methods Prog. Biomed. 140, 93–110 (2017)

41. Zhuang, N.; Zeng, Y.; Tong, L.; Zhang, C.; Zhang, H.; Yan, B. Emotion Recognition from EEG Signals Using Multidimensional Information in EMD Domain. BioMed Res. Int. **2017**, 2017.
42. Li, X.; Yan, J.Z.; Chen, J.H. Channel Division Based Multiple Classifiers Fusion for Emotion Recognition Using EEG Signals. In Proceedings of the 2017 International Conference on Information Science and Technology, Wuhan, China, 24–26 March 2017; Volume 11, p. 07006.
43. Menezes, M.L.R.; Samara, A.; Galway, L.; Sant'Anna, A.; Verikas, A.; Alonso-Fernandez, F.; Wang, H.; Bond, R. Towards emotion recognition for virtual environments: an evaluation of eeg features on benchmark dataset. Pers. Ubiquitous Comput. **2017**, 21, 1003–1013.
44. Brunner C, Billinger M, Seeber M, Mullen TR and Makeig S, (2016), Volume Conduction Influences Scalp-Based Connectivity Estimates. Front. Comput. Neurosci. 10:121. doi: 10.3389/fncom.2016.00121
45. Ioffe, S., & Szegedy, C. (2015). Batch normalization: Accelerating deep network training by reducing internal covariate shift. Paper presented at the International Conference on Machine Learning.
46. D.P. Kingma , J.L. Ba , ADAM: A method for stochastic optimization, 3rd International Conference on Learning Representations, 2015.
47. Bradley MM, Lang PJ (1994) Measuring emotions: the self-assessment manikin and the sematic differential. J Behav Ther Exp Psychiatry 25(1):49–59
48. TensorFlow, 2017. (June 15, 2017).   Retrieved June 25, 2017, from https://www.tensorflow.org/
49. Davidson, R. J., Schwartz, G. E., Saron, C., Bennett, J., and Goleman, D. J. (1979). Frontal versus parietal EEG asymmetry during positive and negative affect. Psychophysiology 16, 202–203.
50. Davidson, R. J., Ekman, P., Saron, C. D., Senulis, J. A., and Friesen, W. V. (1990). Approach-withdrawal and cerebral asymmetry: emotional expression and brain physiology: I. J. Pers. Soc. Psychol. 58, 330–341. doi: 10.1037/0022-3514. 58.2.330
51. Fox, N. A., and Davidson, R. J. (1986). Taste-elicited changes in facial signs of emotion and the asymmetry of brain electrical activity in human newborns. Neuropsychologia 24, 417–422. doi: 10.1016/0028-3932(86)90028-X
52. Davidson, R. J., and Fox, N. A. (1989). Frontal brain asymmetry predicts infants' response to maternal separation. J. Abnorm. Psychol. 98, 127–131. doi: 10.1037/ 0021-843X.98.2.127
53. Tomarken, A. J., Davidson, R. J., and Henriques, J. B. (1990). Resting frontal brain asymmetry predicts affective responses to films. J. Pers. Soc. Psychol. 59, 791–801. doi: 10.1037/0022-3514.59.4.791